\documentclass[a4paper]{llncs}

\usepackage{amssymb}
\usepackage{paralist}
\usepackage{latexsym} 
\usepackage{amssymb,amsmath}
\usepackage{amsmath}
\usepackage[linesnumbered,ruled]{algorithm2e}
\usepackage[]{algorithm2e}
\usepackage{graphics}
\usepackage{lipsum}
\usepackage{graphicx}
\usepackage{graphicx,booktabs,array}
\usepackage{multirow}
\usepackage{tabularx}
\usepackage{array}
\usepackage{listings}
\usepackage{float} 
\usepackage{subfig}
\usepackage{verbatim}
\usepackage{booktabs}

\newcommand{\keywords}[1]{\par\addvspace\baselineskip
\noindent\keywordname\enspace\ignorespaces#1}

\begin{document}

\mainmatter  

\title{Using Swarm Optimization To Enhance  Autoencoder's Images}

\titlerunning{Swarm Optimzation for Enhncing Autoencoders' Images}

%
%
\author{ Maisa Doaud
\and Michael Mayo}
%

\institute{Dept. of Computer Science, Waikato University,\\
 Hamilton, NewZealand\\}

%
%

\toctitle{Lecture Notes in Computer Science}
\tocauthor{Authors' Instructions}
\maketitle

\begin{abstract} 
 Autoencoders learn data representations through reconstruction. Robust training is the key factor affecting the quality of the learned representations and, consequently, the accuracy of the application that use them. Previous works suggested methods for deciding the optimal autoencoder configuration which allows for robust training. Nevertheless, improving the accuracy  of a   trained autoencoder has got limited, if no, attention.  We propose a new approach that  improves the accuracy of a  trained autoencoder's results  and  answers the following question, Given a trained autoencoder, a test image, and using a real-parameter optimizer, can we  generate  better quality reconstructed image version  than  the one generated by the autoencoder?. Our proposed approach  combines both  the decoder part of a trained Resitricted Boltman Machine-based  autoencoder with the Competitive Swarm Optimization algorithm. Experiments show that it is possible to reconstruct images using trained decoder from randomly initialized representations. Results  also show that our approach reconstructed better quality images than the autoencoder in most of the test cases. Indicating that, we can use the  approach  for improving the performance of a pre-trained autoencoder if it does not give satisfactory results.
\end{abstract}

\section{Introduction}

Autoencoders are powerful Neural Network (NN) models that  learn representations of data with multiple levels of abstraction.  Based on the application and the  objective measure,  the abstract   representations  are described as  ``good"  if they are  useful for addressing tasks of interest ~\cite{vincent}. Examples of these tasks are speech recognition,  visual object recognition, image reconstruction and classification. 
\par Learning good  discriminative  
representations using NN-based approaches is very challenging and requires robust training. Factors affecting the   training can be related to the network's configuration, the number  of training examples  \cite{bengio2007greedy} and  the lack of proper data preprocessing (engineering) \cite{bengio2013representation}.  However, deciding the optimal network configuration is  still  determined by  trail-and-error techniques  i.e. by  searching  through a vast space of possible hypermeter combinations.
.
\par  Lots of works have been done so far  to improve the  NNs', including the autoencoders,  accuracy. Most of these  works  concentrated on the networks'  pre-training (data engineering)  and training phases;  so  new network models were suggested/improved and a few parameter selection methods and principles were suggested \cite{ciancio2016heuristic,walczak1999heuristic}.   This paper, in contrast,  presents  a novel approach that improves the  accuracy  of a pre-trained autoencoder results and   answers the following main questions.Without anymore training, is it possible to motivate  a pre-trained autoencoder  to detect more features?.  Given  a pre-trained autoencoder, a test image, and using a real-parameter optimizer, can we generate better quality reconstructed version than  the autoencoder's one?

\par  The approach introduced in this paper  generates good  reconstructed version for  given test images using optimized representations.   We are applying our approach using   a Restricted Boltzmann Machine (RBM)-based  autoencoder and  the Competitive Swarm Optimization (CSO) algorithm.  Only the   decoder part of the trained autoencoder  is  used  in conjunction with  the  (CSO) algorithm  to  generate  optimized representations  that  produce  good  reconstructed versions for the  target test images.

\par 
 Our experiments show that it is possible to reconstruct  interesting images using  the CSO's particle's which are used as inputs to the pre-trained decoder. Moreover, our evaluations proved the efficiency of the proposed approach in  generating good quality reconstructed versions and allowed the decoder to detect some fine details.  
\par The rest of the paper  is organized as follows.  In the next section, we cover background about the image representations and briefly present some related  works.   In the section after that, we describe our proposed new approach. Next, we outline our evaluations. The last  two sections   conclude our work  and present the limitations.

\section{Technical Background}
\subsection{Images Representations and Reconstruction}
 Images can be generally described using  color, shape and texture properties which are extracted using  different techniques such as  Bag of Words (BoW) \cite{bosch:sceneclassification}, Fisher Kernel (FK) \cite{florent:fisherkernels} and  Vector Of Locally Aggregated Descriptors (VLAD) \cite{aggregatinglocal}.  NN's  representations  are other type of representations  that can be  extracted, after training,   by  activating a certain layer or a set of layers and concatenating   \cite{Babenko2014,sharif2014}   or pooling  the results \cite{babenko2015aggregating}. Such  representations are useful for   image classification \cite{leung-Jitendra:representing},visual object  recognition \cite{li:objectbank}, retrieval \cite{philbin:objectretrieval} and  reconstruction \cite{hinton-ruslan:reducing}. 

\par   Images representation and reconstruction are highly related in the NN world.    Dosovitskiyet al,\cite{dosovitskiy2016inverting},  for examples, used a deconvolutional-based  approach   to reconstruct images from  representations learned by a pre-trained deep Convolutional Neural Network (CNN) autoencoders  and proved their efficiency in learning deep images representations. Other  autoencoder-based techniques such as RBM-based autoencoders  \cite{hinton-ruslan:reducing},  discussed in the following subsection, and Stacked Denosing  Autoencoders \cite{vincent}  were also introduced  to extract robust low dimensional  images representations through reconstruction and to reconstruct  test images from low dimensional representations.
\par  Generating good image representations that give satisfactory results  is still a challenge  as it is highly dependent on the robustness of the training process \cite{ciancio2016heuristic}. Neural networks are blackbox predictive tools and designing  them require extensive knowledge engineering and careful parameter and configuration decisions \cite{walczak1999heuristic}.  Walczak and Cerpa \cite{walczak1999heuristic} suggested a set of heuristic principles for designing networks with optimal output performance.  Most recently, Ciancio et al, \cite{ciancio2016heuristic} suggested four different heuristic approaches  to increase the generalization abilities of a neural network. These methods are based, respectively, on the use of genetic algorithms, Taguchi, tabu search and decision trees. The parameters taken into account  are the training algorithm, the number of hidden layers, the number of neurons and the activation function of each hidden layer.
\subsection{RBM-based Autoencoder}
 RBM-based autoencoders were first introduced  by Hinton and Ruslan \cite{hinton-ruslan:reducing} as a non-linear generalization of Principal Component Analysis (PCA). The  network consists of an ``encoder'' part  which  transforms the high-dimensional input data into a low-dimensional representation (the code), and  a ``decoder" part   which recovers the data (image) from the code. 
\par
The autoencoder consists of    two-layer  RBM   network which has stochastic visible and hidden  binary units arranged in sublayers  using symmetrically weighted connections (weights), Figure 1 depicts an illustration of this.
\begin{figure}[htp]
\centering
\includegraphics[width=15em]{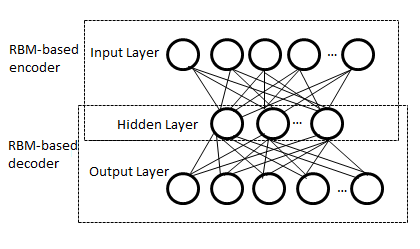}
\caption{RBM autoencoder}
\label{fig:lion}
\end{figure}
\par The  joint configuration of
the visible and hidden units has an energy given by
\begin{equation}
E(v,h)= -\sum_{i\in pixels}b_{i}v_{i} - \sum_{j\in features}b_{j}h_{j} - \sum_{i,j} v_{i}h_{j}w_{ij}
\end{equation}
where $v_{i}$ and $h_{j}$ are the binary states of (visible)  pixel \textit{ i}
and (hidden) feature \textit{ j} respectively;  $b_{i}$ and $ b_{j}$ are their biases; and $ w_{ij}$
is the weight between them.
\par Starting with random weights, the state of  hidden node $h_{j}$   is set to be 1  with a  probability  defined as  : 
\begin{equation}
p(h_{j}=1|v) = \sigma (b_{j}+\sum_{i }v_{i }w_{ij})
\end{equation} 
where $ \sigma(x)$  is the logistic function $1 / [1+exp(-x)]$. 
\par Once the binary states have been chosen for
the hidden nodes, the network sets the visible states by:
\begin{equation}
p(v_{i}=1|h) = \sigma (b_{i}+\sum_{j }h_{j}w_{ij})
\end{equation}
where  $b_{i}$ is the bias of \textit{ i}. 
\par The states of the hidden nodes are then updated once more so
that they represent features of the confabulation. The network   tries iteratively to  minimize  the discrepancy between the input and output data using an optimization algorithm such as   gradient descent and   propagating  the error derivatives through the decoder  and then through the encoder networks  to fine-tune the weights for optimal reconstruction.
\par To use the  RBM with  real-valued images, Ruslan and Hinton ~\cite{ruslan:restrictedboltzmann} suggested pre-training the network  to replace the binary states  by stochastic activities. 
After pretraining, the model unfolds to
produce encoder and decoder networks and fine-tunes the weights  to replace the  stochastic activities
by deterministic, real-valued probabilities. Backpropagation is also used through  the whole autoencoder
to fine-tune the weights for optimal
reconstruction.  
\subsection{Competitive Swarm Optimization (CSO)}

  CSO  \cite{cheng} is fundamentally inspired by Particle Swarm Optimization (PSO) algorithm \cite{kennedy}, which is a conceptually simple optimization algorithm that has attracted considerable research interests so far \cite{suganthan,Kennedy2}. PSO defines a swarm of $n$ particles, each of which has a position and velocity flying in an m-dimensional search space. Each particle evaluates the objective function at its current position and iteratively updates its  velocity and
position  according to  the particle's best position, personal best, and  the entire swarm global  best position found so far. CSO  was suggested to overcome  the  PSO's   poor performance in solving high-dimensional problems and problems with  large number of local optima. 

\par In CSO, the swarm's $n$ particles are randomly allocated   into $\frac{n}{2}$ couples and a competition occurs between the two particles in each
couple. Only the ones with the better fitness, the winners, are passed
 to the next generation (iteration), indexed as $t+1$, of the swarm. Each loser particle  updates
its position and velocity by learning from its winning competitor, and  the updated loser is passed to  generation  $t+1$ after that. Hence, the total number of comparisons occur per generation  is   \( \displaystyle \frac{n}{2} \)  and only the velocity and speed for  \( \displaystyle \frac{n}{2} \) particles are updated per generation. A modified pseudocode of the  CSO (with modifications explained in the next section) is given by Algorithm 1.
\section{Using The CSO For Enhancing  The Autoencoders' Images}
Our main goal is to introduce a novel approach  that improves the accuracy of previously  trained autoencoder results. The idea of the approach is to use the decoder part of a pre-trained autoencoder and a real-parameter optimizer to reconstruct given test images. To apply the approach, we trained an RBM-based autoencoder and  used its decoder with the CSO optimizer. The following subsections presents the methodology and  Figure 2 depicts it.
\subsection{Training The RBM-based Autoencoder} To train an RBM-based autoencoder according to \cite{hinton-ruslan:reducing}, the network has to be configured  by setting the number of  input and output layers' nodes equal to the number of pixels in the training images. Weights and biases are initialized and the network  starts learning  by feeding the training images  through all of its layers to generate output images (reconstructed images) . The discrepancy between the input training image batches and their reconstructed versions are calculated  and   error derivatives are propagated  through the decoder and the encoder parts  to fine-tune the weights and biases. After a finite number  of iterations, the autoencoder will learn the images' features,  and its encoder part will be ready to generate low dimensional  representations for test images that are fed  through its input layer.  These representations  can  be  used as inputs to the decoder part to reconstruct the input test images or used for other applications.

\subsection{Image Reconstruction Using Trained RBM-decoder and CSO}
\par  The second step of the approach represents our main contribution in reconstructing an image using a pre-trained decoder with the CSO algorithm. We will refer to our method as decoder+CSO($m$) to indicate the length $m$ of the used representational vector (Algorithm 1). 
\par Contrary  to studies that discard the decoder part and use the network's  upper part as a fast image dimensionality reduction method,  our novel approach   only uses the  decoder part of the trained autoencoder  and discards the encoder. To generate a reconstructed version $I'$  for  test image $I$ (Equation 4 ),  we  feed  a  swarm $P(t)$ of $n$ randomly initialized  low dimensional vectors ($X_{1}, .., X_{n}$)  through the  trained decoder's layers to get a set of output  images $ I'_{n}$  at its output layer as    depicted by Figure 2 step 2. Note that each particle $X_n$ is  an $m$-dimensional  vector and $m$ equals to the number of nodes in the decoder's first layer i.e. the autoencoder's bottleneck layer.  
\begin{equation}
I'=Decoder(X)
\end{equation}
\begin{figure}[h]
\centering
\includegraphics[width=10cm]{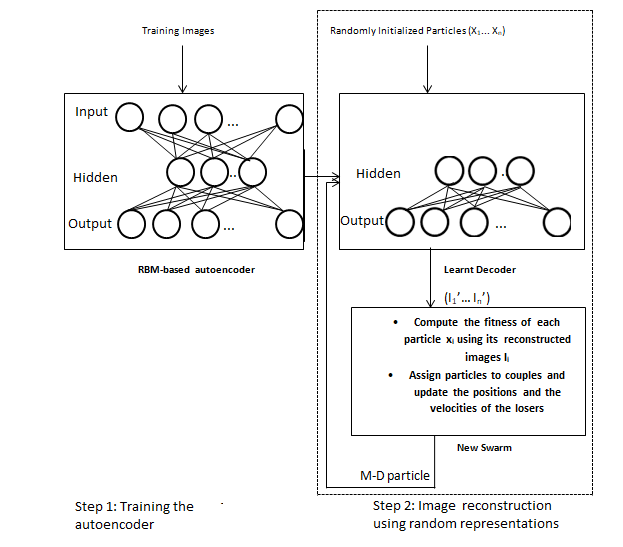}
\caption{RBM-based autoencoder training and image reconstruction  using pre-trained decoder+CSO (m). See Algorithm 1 for better understanding}
\label{fig:lion}
\end{figure} 
\par The   Euclidean norm (given by equation 5) between each  output image $I'_{1..n}$ and  the target test image $I $  is computed   to identify the fittest individuals.
\begin{equation}
F(X)= || (I - I')||_2
\end{equation}
Where $I'$ is the result of decoding representation $X $.

\par 
Particles are randomly allocated into $\frac{n}{2}$ couples and the position and velocity of the   individuals losing each competition are updated  according to their winning partner's existing velocity and the swarm's mean velocity  as shown in  equations 6 \& 7) and cited in \cite{cheng}.
\par
\begin{eqnarray}
 V_{l,k}(t + 1)=&&R_{1}(k,t)V_{l,k}(t) + 
  R_{2}(k,t)(X_{w,k}(t)-X_{l,k}(t))+ \\ 
&& \varphi R_{3}(k,t)(\bar X_{k}(t)-X_{l,k}(t)) \nonumber
\end{eqnarray}

\begin{eqnarray}
    X_{l,k}(t + 1) =&&X_{l,k}(t) + V_{l,k}(t + 1).
\end{eqnarray}
where  $ X_{w,k}$,  $X_{l,k}$,$ V_{w,k}$ and $ V_{l,k}$  are the position and velocity of the winner and the loser  of generation \textit t  respectively, $K \in [1,\frac{n}{2}]$ and $ R_{1}( k,t)$, $ R_{2}( k,t)$, 
$ R_{3}(k, t)$ $\in$
[0,1]\textsuperscript{n} are three randomly
generated vectors after the k-th round of competition  and learning
process in generation \textit t,  $\bar{X}_{k}(t)$ is the mean position value of the
relevant particles and  $\varphi$ is a parameter that controls the influence
of $\bar{X}_{k}$ (t).
\par The process of feeding the particles through the decoder part, calculating the fitness function  and updating each  particle's positions and velocities are continued for a set of iterations to find the best reconstructed version $I' $ for $I$.  

\begin{algorithm}[h]    
 $t= 0$\;
 $count= 0$\;
 $ I= $target test image \;
$  Decoder(X)$, the decoder part of  a pre-trained autoencoder\;
  
 $P(t)$ = { $X_{1},X_{2},..,X_{n}$} randomly initialized  $m$-dimensional  particles\;
    \While{ terminal condition is not satisfied}{
feed each element of  $P(t)$ through the trained decoder to generate $I'_{1}, I'_{2},.., I'_{n}$ using $X_{1},X_{2},..,X_{n}$  according to Equation 4\;
     
   calculate  $F(X)$   between $I$ and $I'_{1}, I'_{2}..I'_{n}$ using Equation 5\; 
   $ U = P(t), O(t + 1) = \O$\;
   \While{$U \neq \O$}{
    randomly remove two particles $X_{rand1}(t)$, $ X_{rand2}(t)$ from U\;

    \eIf{$F(X_{rand1}(t)) \leq F(X_{rand2}(t))$}{
   $X_{w}(t) = X_{rand1}(t), X_{l}(t) = X_{rand2}(t);$
   }{
   $X_{w}(t) = X_{rand2}(t), X_{l}(t) = X_{rand1}(t);$}

    add $ X_{w}(t)$ into $P(t + 1)$\;
 update $ X_{l}(t)$ using (5) and (6)\;
 add the updated $X_{l}(t + 1)$ to $P(t + 1)$\;
 }
  $ t=t+1$;
 
}
\caption{Image Reconstruction  using pre-trained decoder+CSO (m)}
\end{algorithm}
\section*{Experimental Study}
An open-source  java-based deep learning library called  DeepLearning4j (DL4j) (version 0.7.0)  with   the  MNIST and the OlivettiFaces \cite{roweis2012sam}
 datasets  were used to  implement the approach and to perform a set of experiments.  
\par As can be observed by Table 1, the    MNIST dataset consists of 60,000 ($28 \times 28$) training  images for all (0-9) digits. The OlivettiFaces dataset  contains ten ($64\times 64$) images for each of forty different people. We
constructed a training  dataset   by rotating (-90 to 270) and subsampling  images of 30 people to get 10800 ($22 \times 22$) images (i.e. $ 30 \text { people} \times 10 \text { images per person} \times 36 \text { rotations}$). The 500  OlivettiFaces  test images were created in the same way using 10 images of different people.  All training and testing images of both datasets were normalized  using min-max normalizer to get values in [0-1] range. 
\begin{table*}[th]
\centering
  \begin{tabular}{  |c|c|c|c|c|}
  \hline

  Dataset & Dimension  & \begin{tabular}{@{}c@{}}  \# Training \\  Examples  \end{tabular}  &  \begin{tabular}{@{}c@{}}  \# Testing \\  Examples  \end{tabular}   & \begin{tabular}{@{}c@{}}  \ Model\\ Architecture  \end{tabular}\\
 \hline
  MNIST  &  $28 \times 28 $ & 60,000 & 500   &  784-30-784  \\
\hline
   MNIST   &  $28 \times 28 $ &   60,000 & 500  &   784-250-784 \\
\hline
   MNIST   &  $28 \times 28 $ &   60,000 & 500  &   \begin{tabular}{@{}c@{}}7841000-500-256-30\\-256-500-1000-784 \end{tabular}\\

 \hline
 OlivettiFaces  &  $22 \times 22 $ & \begin{tabular}{@{}c@{}}  10,800 (30 \\ people) \end{tabular}  & \begin{tabular}{@{}c@{}} 500 (14 \\ people) \end{tabular}  &   484-30-484  \\
 \hline
 OlivettiFaces  &  $22 \times 22 $ & \begin{tabular}{@{}c@{}}  10,800 (30 \\ people) \end{tabular}  & \begin{tabular}{@{}c@{}} 500 (14 \\ people) \end{tabular}   & 484-300-484   \\
\hline

\end{tabular}
  \caption{Description of the  datasets and network architectures used.}

\end{table*}

\par To test  the efficiency of the proposed approach, we compared the accuracy of  the images reconstructed by  the decoder+CSO(m) (optimized) representations  with the ones reconstructed by the encoder's (non-optimized) representations. The trained encoder part of the RBM-based autoencoder was used to generate a low dimensional representation. This representation is fed through the decoder part to generate a reconstructed version of  the input test image. Hereafter, we will denote  this method by autoencoder($m$) to indicate the length \textit m of the used representational vector.
\par In all test cases, we  set the number of input and output layer's nodes equals to the training  images' size. All  the network's nodes were initialized using DL4j's default   XAVEIR  initialization method where weights  are drawn uniformly in the range $[\frac{-2}{\sqrt{In_{l}+ Out_{l}}},  \frac{2}{\sqrt{In_{l}+Out_{l}}} ]$  such that  `` $In_{l}$"  and ``$Out_{l}$"  are the  number of nodes  sending input to and receiving output from the layer ($l$)  to be initialized  (see  \cite{glorot}). The network  was activated by the sigmoid function, and  back propagated  using the Gradient Descent Optimization algorithm (GDS). 
\par Every  trained decoder was used to reconstruct  500 test  images  using  the CSO's particles.  The swarm   size was set  to 100 randomly  initialized [0-1] individuals  and  the algorithm was run for 100 iterations, which was  good enough to evolve interesting results. We also performed experiments using a smaller swarm size (500)  and  more generations (200),  but results were qualitatively very similar to the ones obtained using 100  particles and 100 iterations, so we are  only presenting the results of this one.

\subsection{Experimental Results and Evaluation}
 
 Five  experiments, three  trained on MNIST dataset and the other two  on the  OlivettiFaces dataset, were performed  using  different network models. The aim  was to test the accuracy of the proposed approach  in reconstructing images from  different datasets and using  different configurations.  

\begin{figure}[h]
   \begin{minipage}{0.48\textwidth}
     \centering
     \includegraphics[width=6 cm]{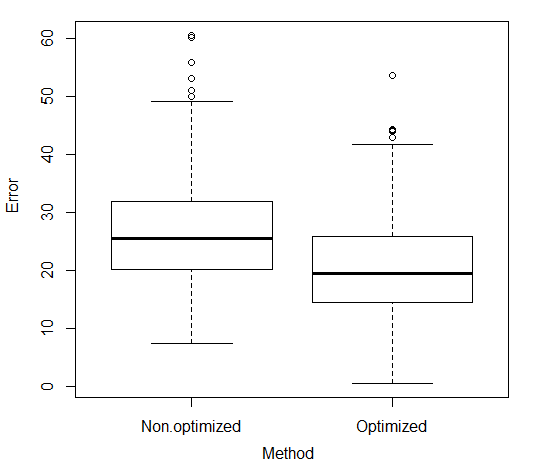}
     \caption{Comparing the performance of autoencoder(30) v.s. decoder+CSO(30) representations in reconstructing MNIST test images. }\label{Fig:Data1}
   \end{minipage}\hfill
   \begin {minipage}{0.48\textwidth}
     \centering
     \includegraphics[width= 6 cm]{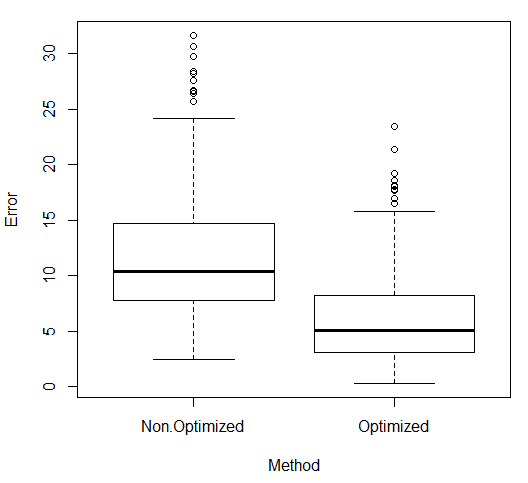}
     \caption{Comparing the performance of autoencoder(250) v.s. decoder+CSO(250) representations in reconstructing  MNIST test images.}\label{Fig:Data2}
   \end{minipage}
\end{figure}
\begin{figure}[!h]
   \begin{minipage}{0.48\textwidth}
     \centering
     \includegraphics[width=6 cm]{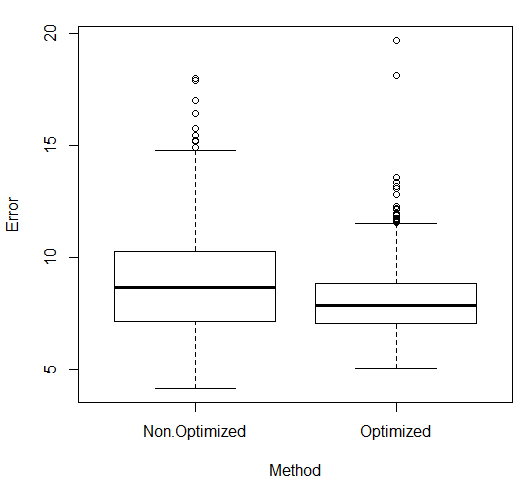}
     \caption{Comparing the performance of autoencoder(30) v.s. decoder+CSO(30) representations in reconstructing   OlivittiFaces test images. }\label{Fig:Data1}
   \end{minipage}\hfill
   \begin {minipage}{0.48\textwidth}
     \centering
     \includegraphics[width=6 cm]{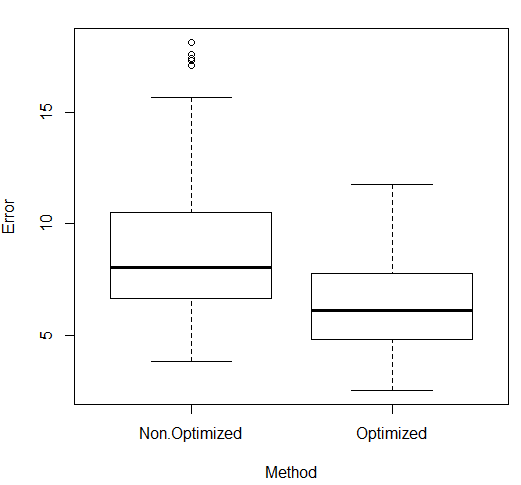}
     \caption{Comparing the performance of autoencoder(300) v.s. decoder+CSO(300) representations in reconstructing   OlivittiFaces test images.}\label{Fig:Data2}
   \end{minipage}
\end{figure}


\par Figures 3-7 depict a set of box-and-whisker plots showing the distribution of the error between target original images and reconstructed versions resulting from pre-trained  autoencoder(m) and decoder+CSO(m).
  Comparing the performance of the  two methods  indicates that the trained decoder+CSO(m) reconstructed better quality images  with remarkably less reconstruction error than the trained encoder(m)  in most test cases. The  decoder+CSO(m) clearly achieved  superior results  with lowest overall  and median reconstruction errors when using three-layered network models for both  MNIST datasets  (Figures 3-4) and  OlivettiFaces dataset (Figure 5-6).    
\par We were also interested in testing our approach using multi layer networks. For fast training, we chose a Multi Layer Percptron (MLP) network  configured using  nine layers with the following  sizes   784-1000-500-256-30-256-500-1000-784. As shown by  Figure 7, Optimization  added only slight little   improvement to the reconstructed images compared to the accuracy  of the multi layer autoencoder images.
\begin{figure}[htp]
\centering
\includegraphics[width=15em]{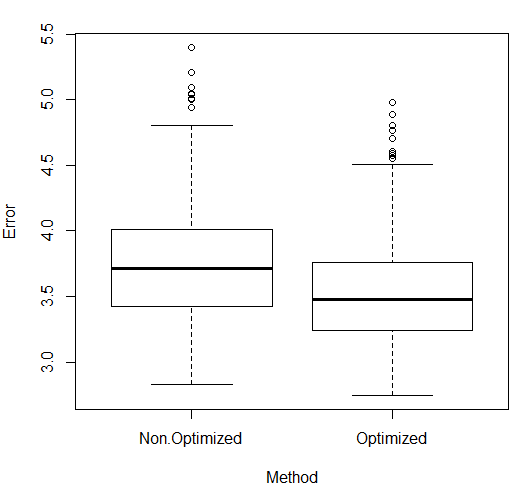}
\caption{Comparing the performance of multi layer autoencoder(30) v.s. Multi layer decoder+CSO(30) representations in reconstructing   MNIST test images.}
\label{fig:lion}
\end{figure}

\subsection*{Qualitative Comparison}

Finally we performed  qualitative comparisons between all related test cases. As shown by  Figures 8-17, the  decoder+CSO(m) method  reconstructed  better quality, less blurry and distorted,  MNIST  images  compared to the autoencoder(m). However, best results obtained, from both methods, when the length of the  representational vector  was 250.

\par Experiments on the OlivettiFaces dataset  show that  the autoencoder(m) method, trained using  our parameter settings,  was able to efficiently learn the orientation of faces but not their fine details (see Figures 12-15). However, using the optimizer  helped in revealing more of the face details.
\par Results generally indicated that using  the CSO  real-parameter optimizer enabled the pre-trained network to detect (filter) more of the test images features.  Indicating that,  the general CSO randomly initialized population was able to activate   some  of the  network's feature detectors  better than the  encoder's representations.

\begin{figure}[h]
   \begin{minipage}{0.48\textwidth}
     \centering
     \includegraphics[width=5 cm]{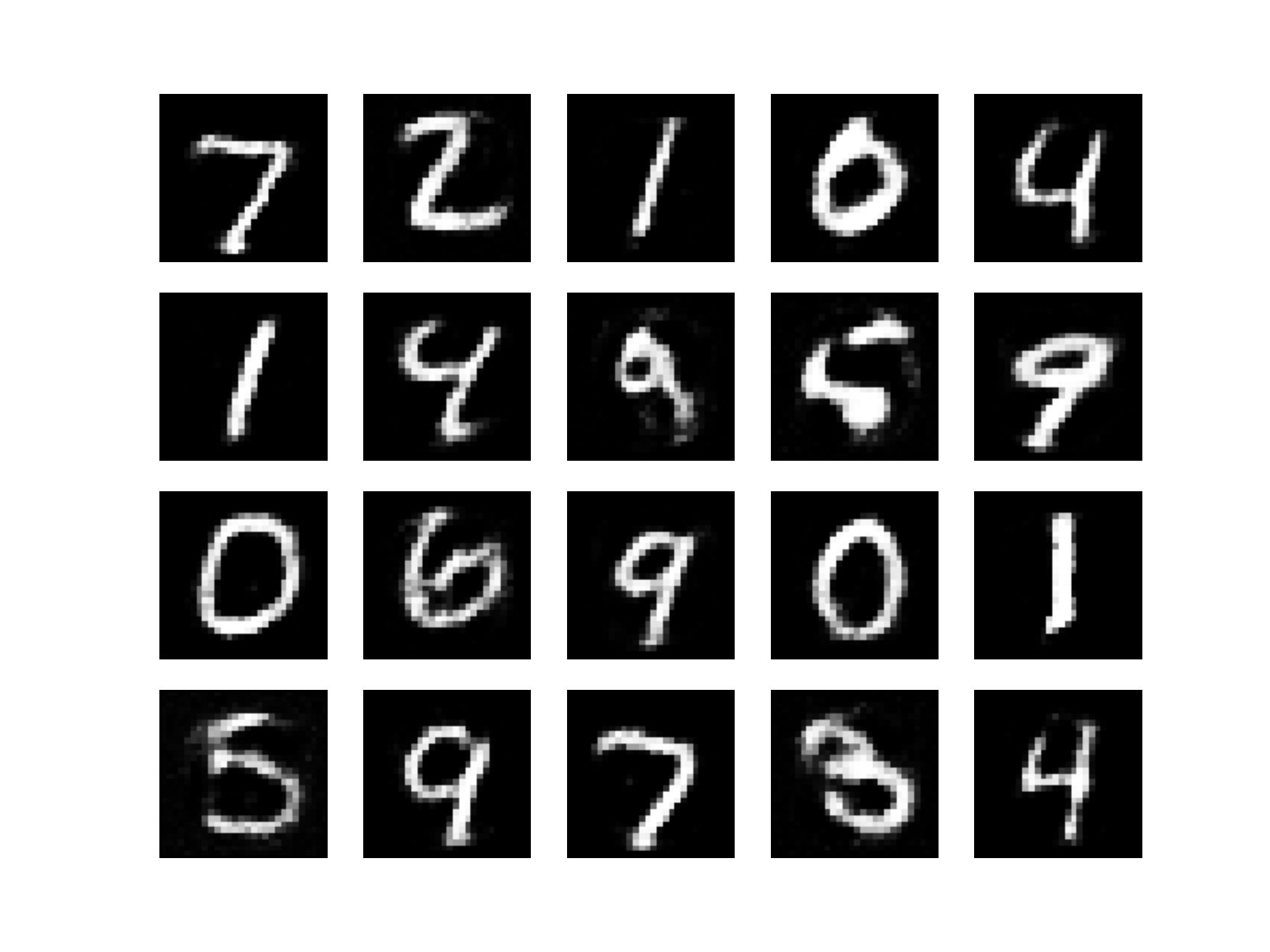}
     \caption{ reconstruction of MNIST using   decoder+CSO(250) representations. }\label{Fig:Data1}
   \end{minipage}\hfill
   \begin {minipage}{0.48\textwidth}
     \centering
     \includegraphics[width= 5 cm]{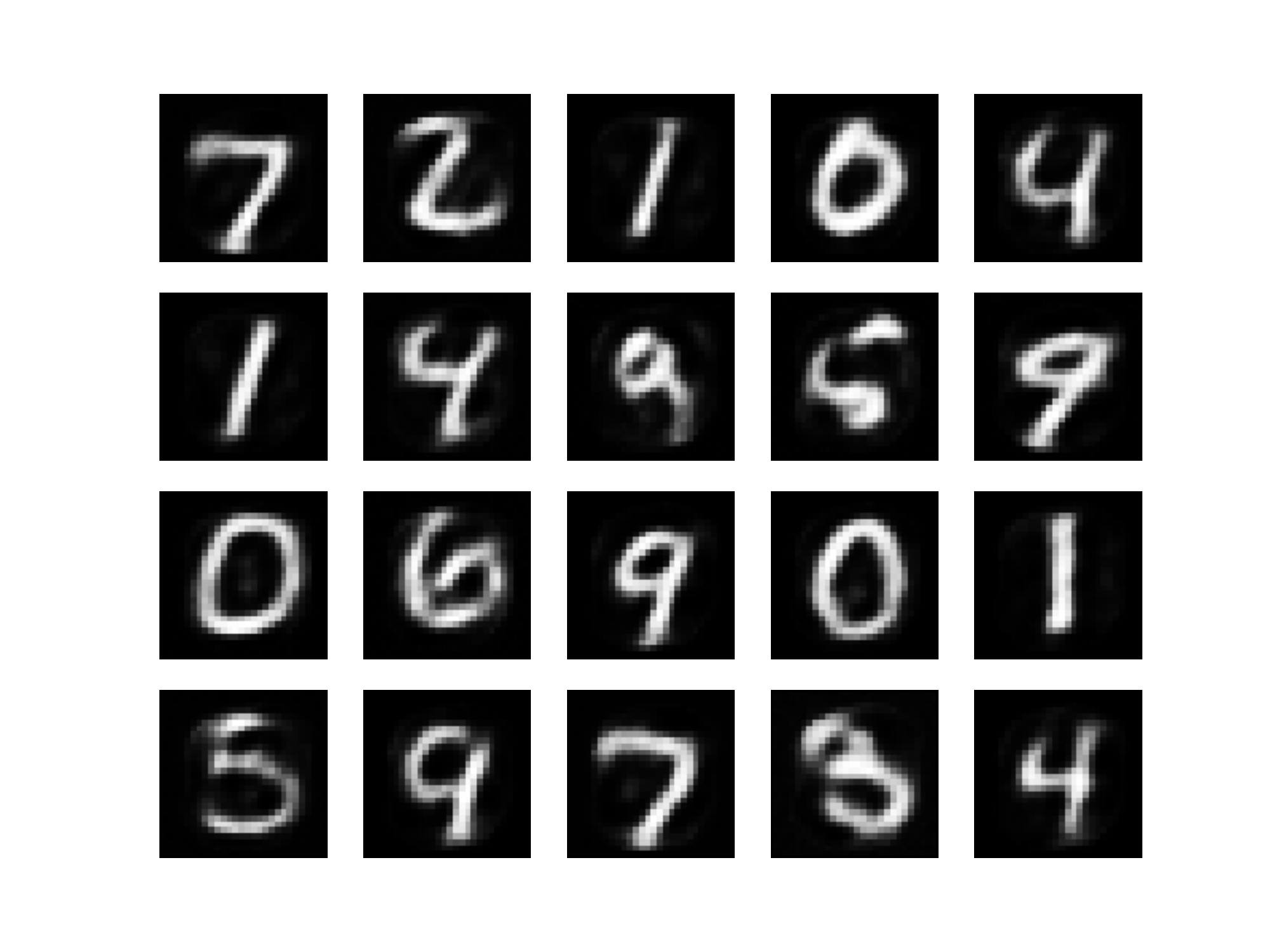}
     \caption{reconstruction of MNIST using   autoencoder(250) representations.}\label{Fig:Data2}
   \end{minipage}
\end{figure}
 \begin{figure}[!htb]
 \begin{minipage}{0.48\textwidth}
     \centering
     \includegraphics[width=5 cm]{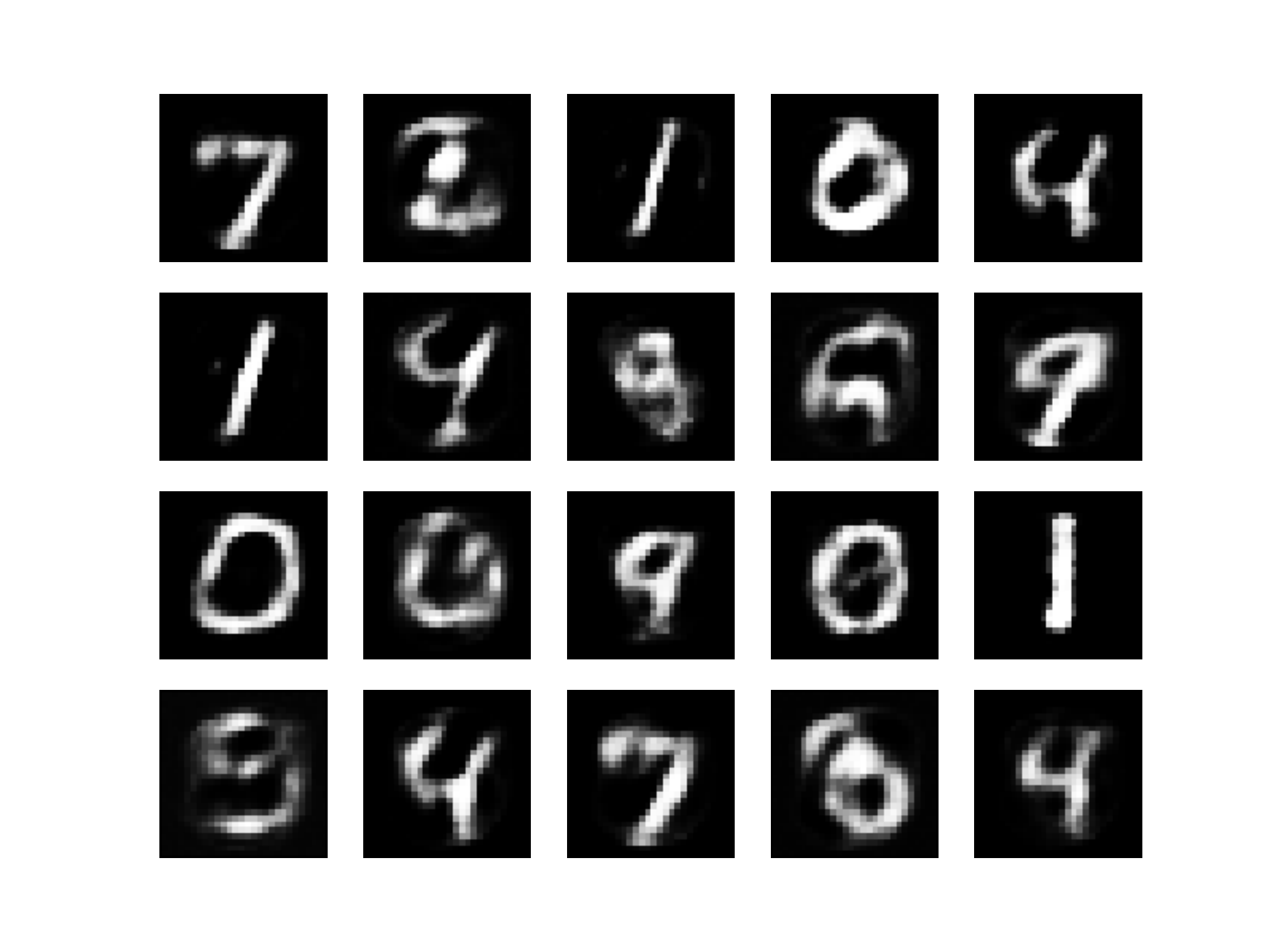}
     \caption{reconstruction of MNIST using    decoder+CSO(30) representations. }\label{Fig:Data1}
   \end{minipage}\hfill
   \begin {minipage}{0.48\textwidth}
     \centering
     \includegraphics[width= 5 cm]{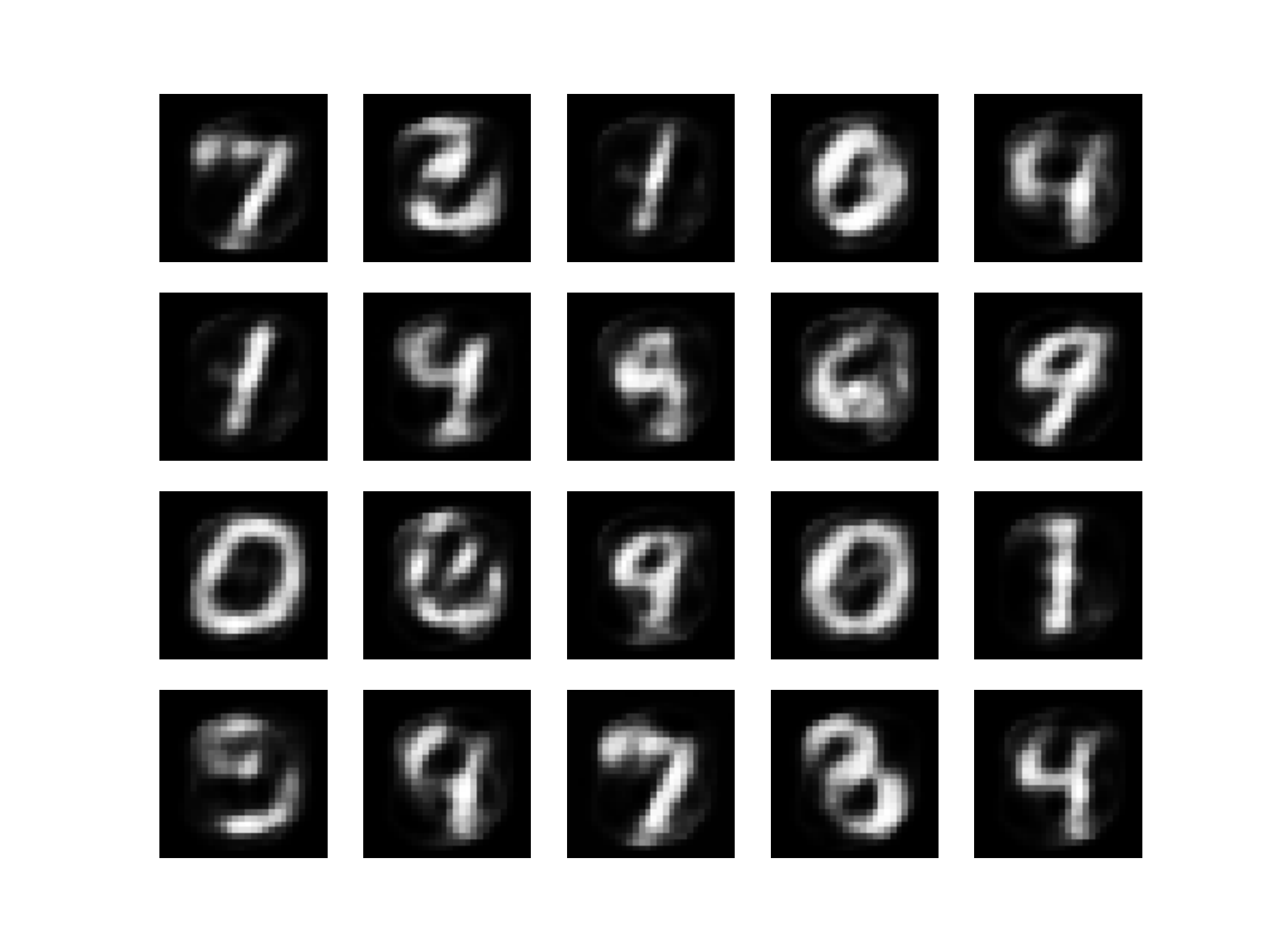}
     \caption{reconstruction of MNIST using   autoencoder(30) representations.}\label{Fig:Data2}
   \end{minipage}
\end{figure}

\begin{figure}[h]
   \begin{minipage}{0.48\textwidth}
     \centering
     \includegraphics[width=5 cm]{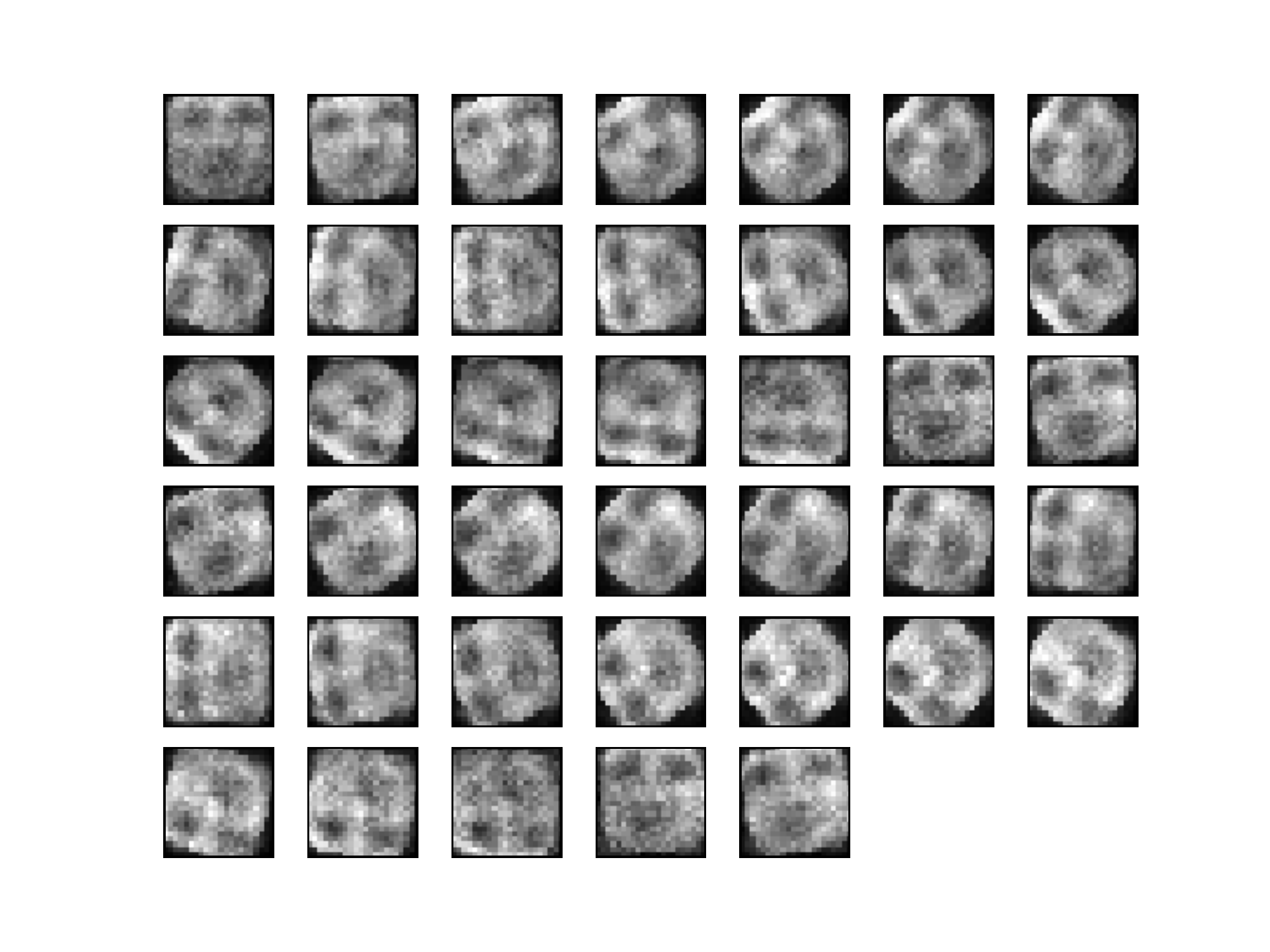}
     \caption{reconstruction of OlivittiFaces using    decoder+CSO(30) representations. }\label{Fig:Data1}
   \end{minipage}\hfill
   \begin {minipage}{0.48\textwidth}
     \centering
     \includegraphics[width= 5cm]{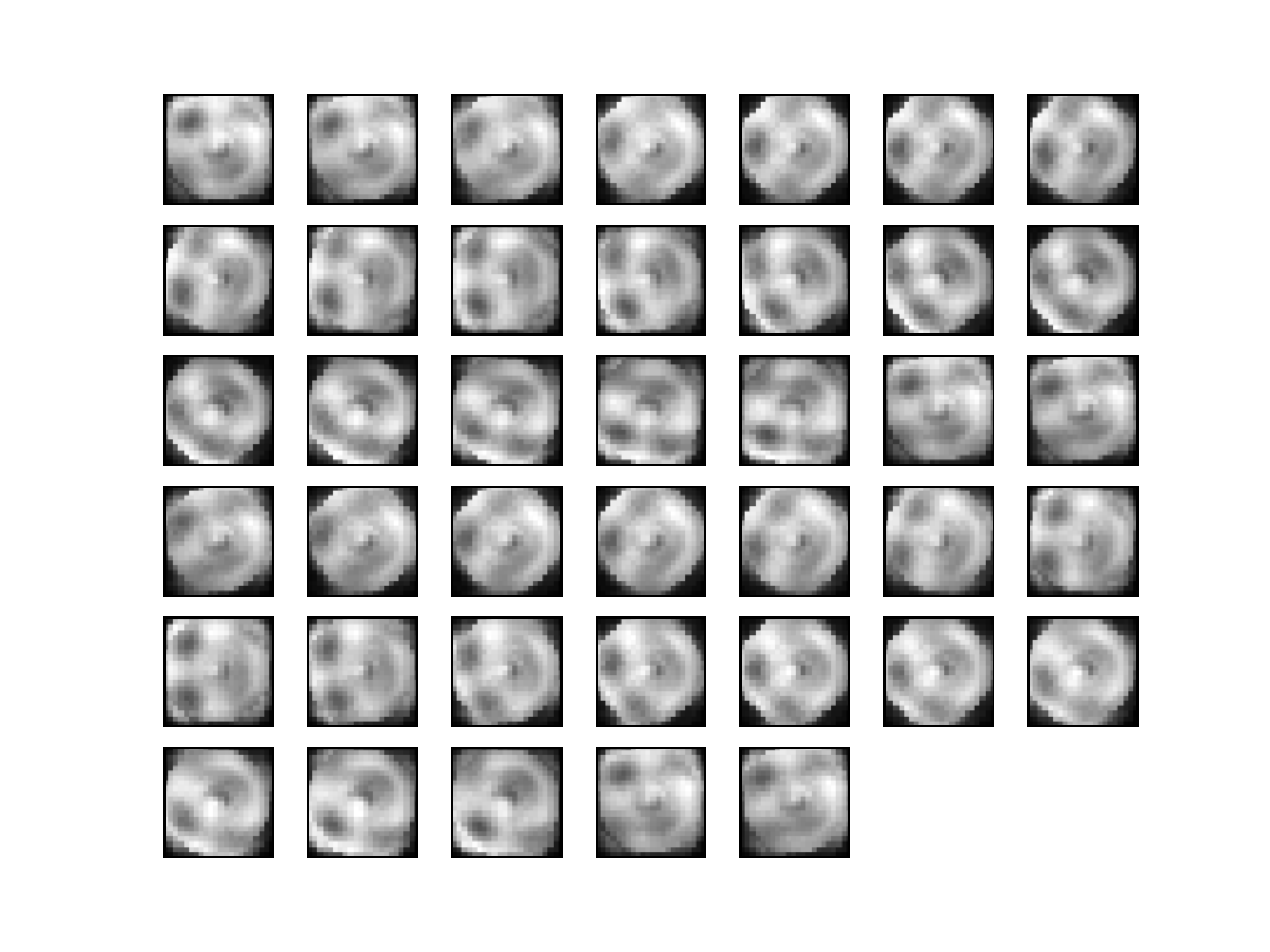}
     \caption{reconstruction of OlivittiFaces using   autoencoder(30) representations.}\label{Fig:Data2}
   \end{minipage}
\end{figure}

\begin{figure}[h]
   \begin{minipage}{0.48\textwidth}
     \centering
     \includegraphics[width=5 cm]{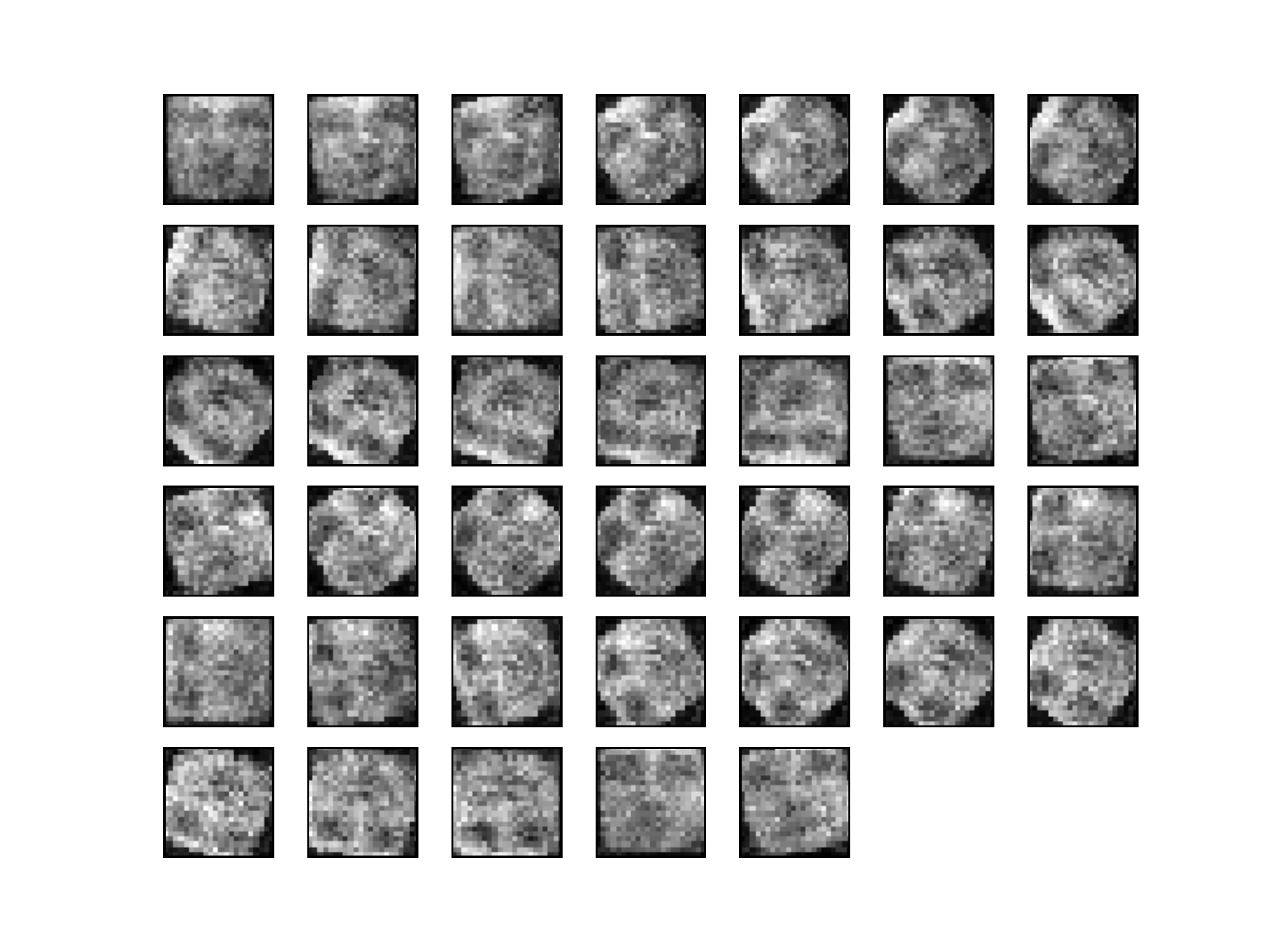}
     \caption{reconstruction of OlivittiFaces using    decoder+CSO(300) representations. }\label{Fig:Data1}
   \end{minipage}\hfill
   \begin {minipage}{0.48\textwidth}
     \centering
     \includegraphics[width= 5 cm]{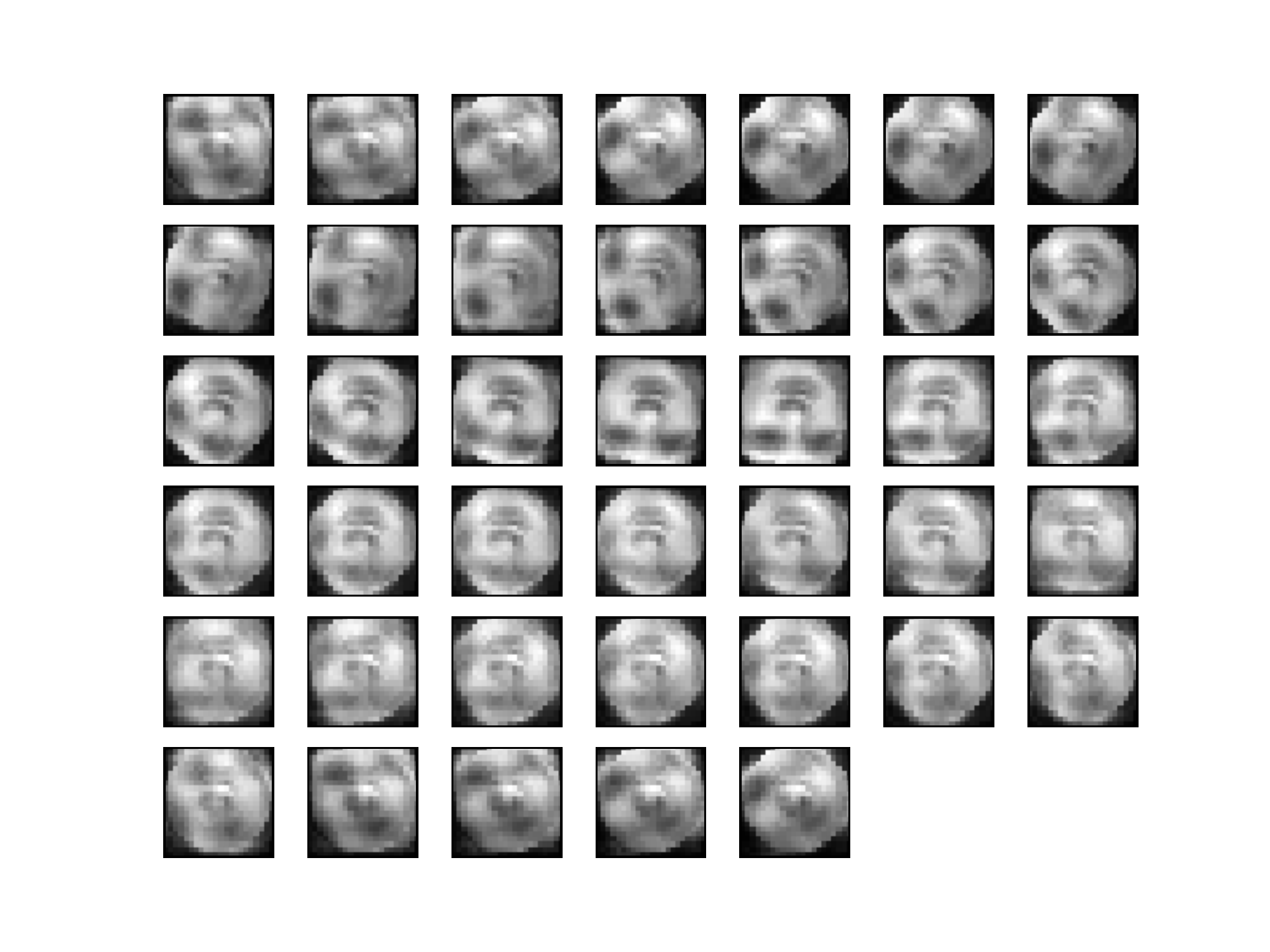}
     \caption{reconstruction of OlivittiFaces using   autoencoder(300) representations.}\label{Fig:Data2}
   \end{minipage}
\end{figure}
\begin{figure}[h]
   \begin{minipage}{0.48\textwidth}
     \centering
     \includegraphics[width=4 cm]{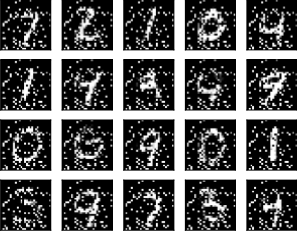}
     \caption{reconstruction of MNIST using Multi layer  decoder+CSO(30) representations. }\label{Fig:Data1}
   \end{minipage}\hfill
   \begin {minipage}{0.48\textwidth}
     \centering
     \includegraphics[width= 4 cm]{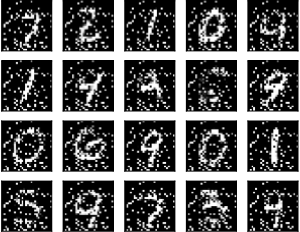}
     \caption{reconstruction of MNIST using Multi layere autoencoder(30) representations.}\label{Fig:Data2}
   \end{minipage}
\end{figure}

\section*{Conclusion}
To conclude, this paper has described a novel  approach  for improving the performance of previously trained autoencoders. The approach was applied by combining  a pre-trained  RBM-based   decoder with the CSO algorithm and was used to reconstruct a set of test  images. Experiments have shown that the suggested approach  is able to reconstruct images using randomly initialized representations. The optimization helped in producing sharper images and detecting finer details so it helped in improving the accuracy of pre-trained autoencoder results.
\par The approach can  be applied  using other types of  autoencoders and/or real-parameter optimizers especially when no satisfactory results obtained from the trained network. In future, we will use the optimized representations in real world  biomedical applications.
\section{Limitations}
Our approach is  proposed  to improve the accuracy of   pre-trained autoencoders results. Reasons  affecting the networks performance can be related to the networks'  learning parameters settings and configuration which are determined by   trials-and-errors. Using the approach with robustly trained or un-trained networks will add no improvement to the results. The approach is helpful when the network's neurons  are  able to detect most of  the images features but more accuracy is required. In this case,  the optimizer, with its  random   population,   could extract more information from the pre-trained network and  increase its  ability in  detecting  fine details and improving the images.

\bibliographystyle{splncs03}
\bibliography{Bib}

\begin{thebibliography}{10}
\providecommand{\url}[1]{\texttt{#1}}
\providecommand{\urlprefix}{URL }

\bibitem{babenko2015aggregating}
Babenko, A., Lempitsky, V.: Aggregating local deep features for image
  retrieval. In: Proceedings of the IEEE International Conference on Computer
  Vision. pp. 1269--1277 (2015)

\bibitem{Babenko2014}
Babenko, A., Slesarev, A., Chigorin, A., Lempitsky, V.: Neural codes for image
  retrieval. In: European Conference on Computer Vision. pp. 584--599. Springer
  (2014)

\bibitem{bengio2013representation}
Bengio, Y., Courville, A., Vincent, P.: Representation learning: A review and
  new perspectives. IEEE transactions on pattern analysis and machine
  intelligence  35(8),  1798--1828 (2013)

\bibitem{bengio2007greedy}
Bengio, Y., Lamblin, P., Popovici, D., Larochelle, H., et~al.: Greedy
  layer-wise training of deep networks. Advances in Neural Information
  Processing Systems  19,  153 (2007)

\bibitem{bosch:sceneclassification}
Bosch, A., Andrew, Z., Muñoz, X.: Scene classification via p\uppercase{LSA}.
  In: European Conference on Computer Vision. pp. 517--530. Springer, Berlin
  Heidelberg (May 2006)

\bibitem{cheng}
Cheng, R., Jin, Y.: A competitive swarm optimizer for large scale optimization.
  IEEE Transactions on Cybernetics (2),  191--204 (February 2015)

\bibitem{ciancio2016heuristic}
Ciancio, C., Ambrogio, G., Gagliardi, F., Musmanno, R.: Heuristic techniques to
  optimize neural network architecture in manufacturing applications. Neural
  Computing and Applications  27(7),  2001--2015 (2016)

\bibitem{dosovitskiy2016inverting}
Dosovitskiy, A., Brox, T.: Inverting visual representations with convolutional
  networks. In: Proceedings of the IEEE Conference on Computer Vision and
  Pattern Recognition. pp. 4829--4837 (2016)

\bibitem{glorot}
Glorot, X., Bengio, Y.: Understanding the difficulty of training deep
  feedforward neural networks. Aistats  9,  249--256 (December 2010)

\bibitem{aggregatinglocal}
Hervé, J., Douze, M., Schmid, C., Pérez, P.: Aggregating local descriptors
  into a compact image representation. In: Computer Vision and Pattern
  Recognition (CVPR). pp. 3304--3311. IEEE Conference (June 2006)

\bibitem{hinton-ruslan:reducing}
Hinton, G., Ruslan, S.R.: Reducing the dimensionality of data with neural
  networks. Science  313(5786),  504--507 (July 2006)

\bibitem{kennedy}
Kennedy, J., Eberhart, R.: Particle swarm optimization. IEEE International
  Conference on Neural Networks (5786),  504--507 (1995)

\bibitem{Kennedy2}
Kennedy, J., Mendes, R.: Population structure and particle swarm performance.
  In: Evolutionary Computation, 2002. CEC'02. Proceedings of the 2002 Congress
  on 2002. vol.~2, pp. 1931–--1938. IEEE (2002)

\bibitem{leung-Jitendra:representing}
Leung, T., Jitendra, M.: Representing and recognizing the visual appearance of
  materials using three-dimensional textons. International Journal of Computer
  Vision  43(1) (June 2001)

\bibitem{li:objectbank}
Li, L.J., Su, H., Lim, Y., Fei-Fei., L.: Object bank: An object-level image
  representation for high-level visual recognition. International Journal of
  Computer Vision  107(1) (March 2014)

\bibitem{florent:fisherkernels}
Perronnin, F., Christopher, D.: Fisher kernels on visual vocabularies for image
  categorization. In: Computer Vision and Pattern Recognition. pp. 1--8. IEEE
  Conference (June 2007)

\bibitem{philbin:objectretrieval}
Philbin, J., Ondrej, C., Isard, M., Sivic, J., Zisserman, A.: Object retrieval
  with large vocabularies and fast spatial matching. In: 2007 IEEE Conference
  on Computer Vision and Pattern Recognition. IEEE (June 2007)

\bibitem{roweis2012sam}
Roweis, S.: sam roweis: data.[www page]. URL http://www. cs. nyu. edu/\~{}
  roweis/data. html  (2012)

\bibitem{ruslan:restrictedboltzmann}
Ruslan, S., Mnih, A., Hinton, G.: Restricted boltzmann machines for
  collaborative filtering. In: Proceedings of the 24th international conference
  on machine learning. pp. 791--798. ACM (June 2007)

\bibitem{sharif2014}
Sharif~Razavian, A., Azizpour, H., Sullivan, J., Carlsson, S.: {CNN} features
  off-the-shelf: an astounding baseline for recognition. In: Proceedings of the
  IEEE Conference on Computer Vision and Pattern Recognition Workshops. pp.
  806--813 (2014)

\bibitem{suganthan}
Suganthan, P.N.: Particle swarm optimiser with neighbourhood operator. In: CEC
  99. Proceedings of IEEE Congress on Evolutionary Computation. pp.
  1931–--1938. IEEE (1999)

\bibitem{vincent}
Vincent, P., Larochelle, H., Lajoie, I., Bengio, Y., Manzago, P.A.: Stacked
  denoising autoencoders: Learning useful representations in a deep network
  with a local denoising criterion. Journal of Machine Learning Research  11,
  3371--3408 (December 2010)

\bibitem{walczak1999heuristic}
Walczak, S., Cerpa, N.: Heuristic principles for the design of artificial
  neural networks. Information and software technology  41(2),  107--117 (1999)

\end{thebibliography}

\end{document}